\documentclass[conference]{IEEEtran}
\IEEEoverridecommandlockouts
\usepackage{cite}
\usepackage{amsmath,amssymb,amsfonts}
\usepackage{algorithmic}
\usepackage{graphicx}
\usepackage{textcomp}
\usepackage{xcolor}
\usepackage{multirow}
\usepackage{multicol}
\usepackage{booktabs}
\usepackage{enumitem}
\usepackage{amsfonts}
\usepackage{symbols}
\usepackage{hyperref}

\newcommand*{\Scale}[2][4]{\scalebox{#1}{$#2$}}%

\def\BibTeX{{\rm B\kern-.05em{\sc i\kern-.025em b}\kern-.08em
    T\kern-.1667em\lower.7ex\hbox{E}\kern-.125emX}}

\begin{document}

\title{Active Semantic Mapping with Mobile Manipulator in Horticultural Environments\\
}

\author{Jose Cuaran$^{1}$, Kulbir Singh Ahluwalia$^{1}$, Kendall Koe$^{1}$, Naveen Kumar Uppalapati$^{3}$, and Girish Chowdhary$^{1,2}$

\thanks{The authors are with (1) the Department of Computer Science, (2) the Department of Agricultural and Biological Engineering  and (3) National Center for Supercomputing Applications at University of Illinois, Urbana-Champaign.
        }%
\thanks{{Correspondence to \tt\small \{jrc9,girishc\}@illinois.edu}}
}

\maketitle

\begin{abstract}
Semantic maps are fundamental for robotics tasks such as navigation and manipulation. They also enable yield prediction and phenotyping in agricultural settings. In this paper, we introduce an efficient and scalable approach for active semantic mapping in horticultural environments, employing a mobile robot manipulator equipped with an RGB-D camera. Our method leverages probabilistic semantic maps to detect semantic targets, generate candidate viewpoints, and compute corresponding information gain. We present an efficient ray-casting strategy and a novel information utility function that accounts for both semantics and occlusions. The proposed approach reduces total runtime by 8\% compared to previous baselines. Furthermore, our information metric surpasses other metrics in reducing multi-class entropy and improving surface coverage, particularly in the presence of segmentation noise. Real-world experiments validate our method's effectiveness but also reveal challenges such as depth sensor noise and varying environmental conditions, requiring further research.\\
\href{https://github.com/jrcuaranv/nbv_planning}{https://github.com/jrcuaranv/nbv\_planning}.

\end{abstract}

\vspace{-0.1cm}
\section{Introduction}

Semantic maps in agricultural environments provide robots with crucial information to guide actions, like navigating rows or harvesting fruit. It also supplies data to farmers or management systems for tasks such as predicting yields and monitoring growing rates \cite{kootstra2024advances}. However, building these maps presents several challenges, such as variations in environmental conditions, wind disturbances, and incomplete observations caused by occlusions.

Various works address the problem of mapping in agricultural environments \cite{smitt2023pag, pan2023panoptic, dong2020semantic}. Most of these works use fixed cameras to collect images, which are post-processed with approaches like Structure from Motion or Neural Radiance Fields to achieve a 3D reconstruction. Such approaches often suffer from limited viewpoints and self-occlusions that are common in agricultural scenes. Moreover, for tasks like yield prediction, semantics corresponding to fruits are more relevant than semantics corresponding to leaves or other parts, making target-aware mapping necessary.

\begin{figure}[htbp]
\centerline{\includegraphics[width=90mm]{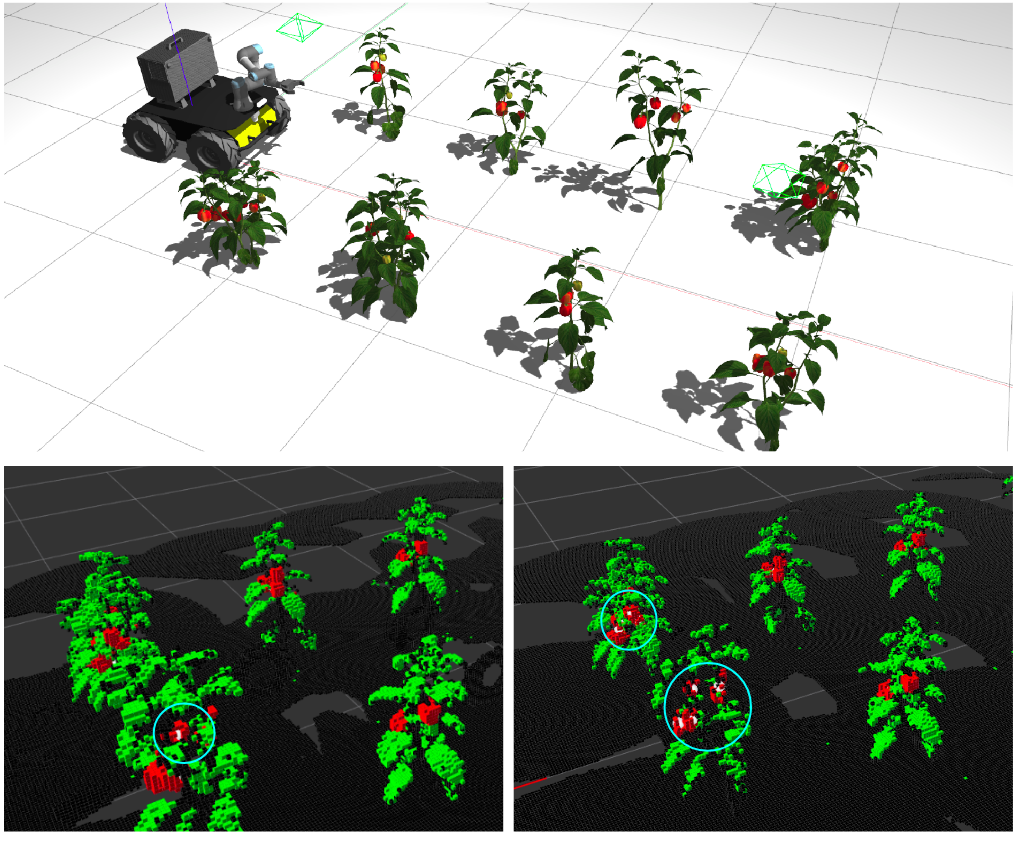}}
\caption{Top: Simulation environment with capsicum plants and Husky Robot. Bottom left: Reconstructions done using our approach. Bottom right: Reconstruction done using predefined dense scanning. Blue circles highlight incomplete fruit areas.}
\label{reconstruction}
\vspace{-0.7cm}
\end{figure} 

Thus, active mapping approaches, common in other environments, have been adopted to improve the reconstruction quality of agricultural environments. A popular method is Next Best View (NBV) planning, which relies on computing a utility value for potential viewpoint candidates \cite{burusa2022attention, zaenker2021viewpoint, menon2023nbv}. Although prior works show promising results in building target-aware reconstructions, some works still rely on the assumption that bounding boxes of targets are given \cite{burusa2022attention}. In addition, binary occupancy maps are commonly used, which only provide information about the occupancy but not the probability of semantic classes. Finally, most of these studies rely on ray-casting, a technique that involves tracing the paths of multiple rays from a sensor to gather occupancy information for each voxel. This method can be computationally expensive if not implemented efficiently.

To address these limitations, we present an efficient and scalable NBV approach for active semantic mapping in horticultural environments. Unlike prior works based on binary occupancy maps \cite{zaenker2021viewpoint,burusa2022attention,menon2023nbv}, we leverage semantic probabilistic maps \cite{asgharivaskasi2023semantic} to directly detect the semantic targets without the need for predefined bounding boxes. We use these targets to sample multiple viewpoint candidates around them. An information utility value is computed for each candidate for which an efficient ray-tracing strategy is implemented. Finally, we evaluate the performance of our method on different Information Gain (IG) metrics, and propose a new metric that leverages the multi-class probability map.

In summary, the main contributions of this paper are:
\begin{itemize}
    \item An approach for target-aware semantic mapping in horticultural environments leveraging state-of-the-art multi-class probabilistic maps. 
    \item A simple but effective strategy to make ray casting faster and enable the efficient evaluation of viewpoint candidates.
    \item A novel information utility function that takes into account occlusions, proximity to targets, and semantics.
    \item An evaluation of the performance of our method and utility function considering the segmentation noise, which is common in agricultural environments because of environment variations, but rarely considered in previous works.
\end{itemize}

\begin{figure*}[htbp]
\centerline{\includegraphics[width=180mm]{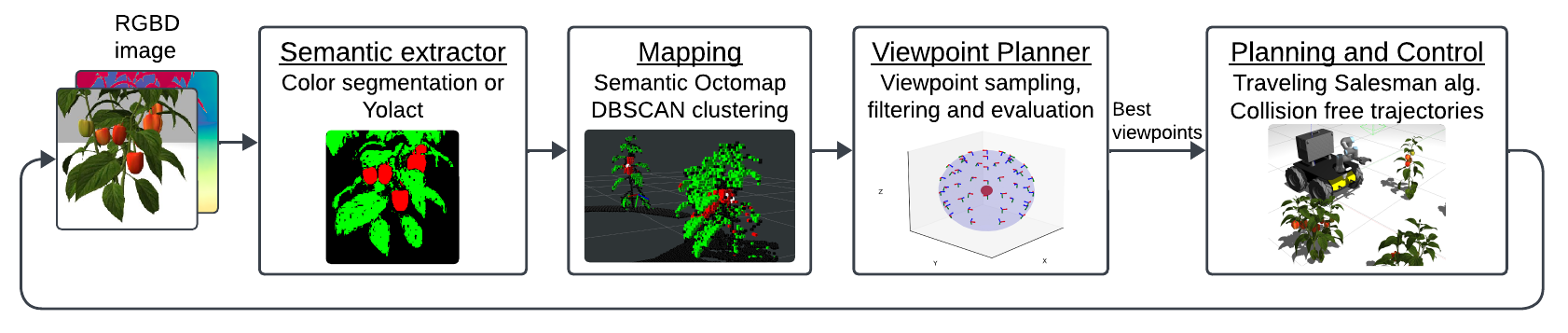}}
\vspace{-0.2cm}
\caption{System overview. Semantic extractor: semantic segmentation is used to extract fruits, leaves, and background in the scene. Mapping: a 3D semantic and probabilistic representation of the environment is generated. Viewpoint planner: It generates potential viewpoint candidates and chooses those with the highest information gain. Planning and control:  collision-free trajectories to the desired goals and control signals are generated.}
\vspace{-0.1cm}
\label{system_overview}
\vspace{-14pt}
\end{figure*}   

\vspace{-0.1cm}
\section{Related works}
\label{sec:relatedworks}
Numerous studies have addressed the problem of mapping and 3D reconstruction in agricultural environments \cite{gao2021canopy,dong20174d, dong2020semantic, nellithimaru2019rols}. However, the majority of these approaches are passive, where sensors, such as cameras or LiDAR, are fixed on a robot or handheld while capturing a sequence of images or scans. The reconstructed 3D point clouds are subsequently used to estimate parameters such as canopy volume, trunk diameter, tree height, and fruit count. While these methods are effective in capturing details at the plant level, they are less suitable for tasks requiring the estimation of the shape, volume, or pose of individual fruits, as they are often affected by occlusions and limited viewpoints. Although fruit counting is performed in \cite{dong2020semantic} and \cite{nellithimaru2019rols}, these analyses are conducted in image space, thereby neglecting occluded fruits that can only be captured using 3D models.

Active mapping approaches have been increasingly proposed to enhance the quality of reconstructions in agricultural environments. Among these approaches, Next Best View (NBV) planning is one of the most common techniques. NBV planning involves identifying sensor poses or trajectories that maximize the information gain (IG), thereby reducing map uncertainty while minimizing associated costs, such as path length or time \cite{burusa2022attention, zaenker2021viewpoint, menon2023nbv}. For example, Burusa \textit{et al.} \cite{burusa2022attention} proposes a method for the active 3D reconstruction of tomato plants using a robotic manipulator. In this approach, the selection of the next view is based on the IG calculated for random viewpoint candidates, leveraging a probabilistic occupancy map of the environment. However, a notable limitation of this work is its reliance on predefined 3D bounding boxes for various parts of the plants, which guides the active planning algorithm toward the desired semantics. In contrast, our method addresses this limitation by maintaining a semantic map of the environment, which eliminates the need for predefined bounding boxes.

Zaenker \textit{et al.} \cite{zaenker2021viewpoint} introduce an NBV planning approach aimed at mapping sweet pepper plants and estimating the size and position of fruits. Their method employs a probabilistic octree representation of the environment, where each cell encodes occupancy and region of interest (ROI) probabilities. To direct the algorithm's focus toward fruit areas, they generate random viewpoint candidates by leveraging frontier nodes near ROI regions. General frontier nodes are also utilized to encourage exploration and the discovery of new ROIs. The authors report improvements in volume coverage and accuracy compared to pure frontier-based exploration that does not account for semantic ROIs. Building upon this work, our approach introduces several enhancements. Unlike their method, which maintains a single ROI probability for each cell, we maintain a multi-class probability distribution, allowing us to handle multiple semantic classes. Additionally, while their approach samples random candidate viewpoints for each frontier node near the ROIs, potentially leading to redundant viewpoints or the omission of critical ones, our method uniformly samples potential viewpoints around fruit clusters. This ensures uniform coverage and also more efficiency, as the number of clusters is less than the number of frontier voxels.

Recent studies have proposed NBV planning approaches that do not rely exclusively on ray casting, which is known to be effective but computationally intensive. For instance, \cite{menon2023nbv} presents a method that utilizes shape completion with superellipsoids to identify optimal viewpoints. By leveraging the predicted surfaces of fruits, this approach guides the sensor toward viewpoints targeting missing surface regions. While this method significantly reduces planning time compared to ray-casting-based techniques, its application is limited to certain types of crops. It is also dependent on the accuracy of the shape completion algorithm, which continues to be an area of active research. Another notable contribution in this line of research is 3D Move-to-see \cite{lehnert20193d}. This method employs a custom 3D camera array that optimizes a utility function through gradient ascent. A single shot is sufficient to guide the camera toward a viewpoint that maximizes the visible fruit area while avoiding occlusions. However, this method addresses only the local path planning problem and must be integrated with a global path planning strategy to achieve effective active mapping \cite{zaenker2021combining}.  


Octomap \cite{hornung2013octomap} has been the primary representation used in most prior works due to its scalability and ability to encode probabilistic information. These maps, which contain binary occupancy probabilities for each cell, enable the computation of various information gain metrics by counting the number of unknown voxels or calculating entropy. Some proposed metrics also consider factors such as visibility and proximity to specific targets \cite{delmerico2018comparison}. Recently, \cite{asgharivaskasi2023semantic} extended this framework to multi-class probability maps, where each voxel, in addition to the binary occupancy probability, includes a categorical distribution over multiple semantic classes. Building on this framework, we propose an information gain metric that incorporates occlusions, proximity, and class probability to guide the algorithm in prioritizing certain targets.

\vspace{-0.1cm}
\section{Methods}
\label{sec:methods}
We aim to create a semantic and geometric representation of plants in a horticultural environment, focusing on specific targets such as fruits with an active mapping approach. We assume that the plants are distributed along rows so that a mobile manipulator can go through the middle of them collecting RGBD images while building the map. The plants' height is also assumed to be known. 

Fig. \ref{system_overview} presents an overview of our system, which is composed of four modules: (i) Semantic Extractor, (ii) Mapping, (iii) Viewpoint Planner, (iv) Planning and Control. Given an RGBD observation of the scene, the semantic extractor generates masks for $K$ semantic classes. The depth images and masks are merged into a semantic point cloud, which is then used by the mapping module. This module creates a probabilistic semantic map encoded as an octree data structure, where each voxel stores occupancy and semantic probabilities. Voxels with a high probability of belonging to a desired target class $k_d$ are clustered into different groups. Each cluster is then utilized by the viewpoint planning module to generate a set of candidate viewpoints, which are subsequently filtered and evaluated using a utility function. The viewpoints with the highest utility values are finally executed. Further details are given below.

\noindent\textbf{(i) Semantic extractor:}
We define fruits, leaves, and background as our three semantic classes, with fruits as the main semantic class. Given an RGB image, semantic segmentation is applied to extract pixel labels corresponding to those classes. We perform color segmentation based on HSI color values for simulation, whereas for real-world experiments we use the Yolact model \cite{bolya2019yolact}.  

\noindent\textbf{(ii) Mapping:}
We use Semantic Octomap \cite{asgharivaskasi2023semantic} to obtain an efficient, compact, and scalable map representation. Given a semantic point cloud and camera pose, Semantic Octomap creates a probabilistic semantic octree where each voxel $x$ encodes a categorical distribution $p_c(x)$ over different classes as well as a binary occupancy probability value $p_o(x)$.  
As we know the height of the plants and the robot location with respect to the crop rows, we execute a sparse scanning with predefined camera poses around the plants. With these images, Semantic Octomap creates an initial and incomplete semantic octree.  Subsequently, the voxels classified as fruits according to the semantic probability values $p_c(x)$ are used to compute fruit clusters applying the DBSCAN clustering algorithm\cite{dbscan_clustering_algo_paper}. We set the neighborhood threshold $\epsilon$ equal to the expected average fruit radius, ensuring that fruits close to each other are merged into a single cluster, thus preventing redundant viewpoints when sampling viewpoint candidates. This parameter, along with other parameters described in this paper are summarized in Table \ref{tab:parameters}.


\noindent\textbf{(iii) Viewpoint Planner:}
We present an NBV planner that refines an incomplete map by selecting the best viewpoints to improve reconstruction, focusing on specific semantic classes. For each fruit cluster centroid, we sample viewpoints on a sphere with radius $r$, using $N_{\theta}$ elevation and $N_{\phi}$ azimuth angles in the ranges $[30^\circ, 150^\circ]$ and $[0^\circ, 360^\circ]$, respectively. Viewpoints near the manipulator's workspace are filtered. Subsequently, a utility value is computed for each candidate viewpoint based on the IG for selecting the top-k utility viewpoints. These selected high IG viewpoints are then passed onto the MoveIt motion planner.


\noindent\textbf{Filtering out candidate viewpoints.} 
Sampling viewpoints uniformly around each cluster centroid generates many candidates, necessitating a filtering strategy to reduce computational overhead before viewpoint evaluation. We compute points defining the manipulator's workspace using forward kinematics for multiple joint combinations and move candidate viewpoints into the arm workspace. Viewpoints whose viewing direction does not intersect the workspace are discarded, with intersections computed via the Nearest Neighbor algorithm.

\noindent\textbf{OSAMCEP - A novel IG metric.}
Leveraging the multi-class probability information of our semantic map, we propose an information metric called Occlusion and Semantic Aware Multi-Class Entropy with Proximity Count (OSAMCEP), which considers occlusions and focuses on desired semantic targets. For a viewpoint $v$ with a set of cast rays $\mathcal{R}_v$, a set $\mathcal{X}$ of observed voxels per ray and the information utility function ${I(x)}$, we compute the IG, $\mathcal{G}_v$, as: 

\vspace{-0.3cm} 
\begin{equation}
\Scale[0.95]
{
\mathcal{G}_v = \sum_{r \in \mathcal{R}_v}\sum_{x \in \mathcal{X}}I(x)
\label{eq:osamcep}
}
\end{equation}
\vspace{-0.5cm}

Given a voxel $x$ with a probability of occupancy $p_o(x)$, and $p_c(x)$ the categorical distribution over $K$ semantic classes, the corresponding information utility function  ${I(x)}$ is computed as seen in Equation \ref{eq:I_x_calculation}:

\vspace{-0.5cm} 
\begin{equation}
\Scale[0.85]
{I(x) =
\begin{cases}
    P_v(x)H(x) & \text{,if semantic class } l_x \text{ is unknown or}\\
    & \text{semantic target and } dist(x) < max\_dist \\
    0  & \text{,otherwise}
\end{cases}
\label{eq:I_x_calculation}
}
\end{equation}
\vspace{-0.3cm}

\vspace{-0.4cm} 
\begin{equation}
\Scale[0.95]
{
    P_v(x_n) = \prod_{i=1}^{n-1}(1-p_o(x_i))
    \label{eq:voxel_visible_prob}
}
\end{equation}
\vspace{-0.5cm} 

\vspace{-0.4cm} 
\begin{equation}
\Scale[0.95]
{
    H(x) = -\sum_{k=0}^{K}p_c(l_x=k)\ln p_c(l_x=k)
    \label{eq:h_x}
}
\end{equation}
\vspace{-0.5cm}

Where $P_v(x_n)$ is the probability of voxel $x_n$ being visible considering the previous voxels $x_i$ traversed on the ray before reaching voxel $x_n$ as seen in Equation \ref{eq:voxel_visible_prob}.
$H(x)$ is the multi-class entropy defined in Equation \ref{eq:h_x}. $dist(x)$ is the distance between a semantic target and voxel $x$, and $max\_dist$ is a threshold distance value for determining the relevant voxels close to the semantic target.

Unlike other metrics proposed in previous works like \cite{zaenker2021viewpoint, burusa2022attention} which rely only on unknown voxels in proximities to the targets to compute the IG, we consider both unknown and semantic voxels. This encourages the algorithm to reduce the entropy of specific semantic nodes, making target-aware mapping more effective. In addition, while previous works \cite{zaenker2021viewpoint, burusa2022attention} determine the presence of occlusions based on occupancy thresholds, we employ the probability of visibility $P_v(x)$ suggested in \cite{delmerico2018comparison}, computed along each ray, resulting in a more consistent estimation for occlusions.


\noindent\textbf{Increasing Ray Tracing Efficiency.}
Ray-tracing all image pixels to compute each viewpoint's utility is computationally expensive, so uniform downsampling is often used to decrease runtime. However, relevant semantic regions can be missed depending on the distance from the scene because of downsampling. To address this, we propose two strategies. First, downsample pixels with a step size $ds$ inversely proportional to the target distance $z$:

\vspace{-0.2cm} 
\begin{equation}
\Scale[1.2]
{
    ds = \frac{\delta S * F_x}{z}
    \label{eq:downsampling}
}
\end{equation}
\vspace{-0.5cm}


where $\delta S$ is the octomap resolution and $F_x$ the camera focal length. The second strategy assumes distant rays do not impact the current cluster's IG and discards them during ray tracing. Using the typical fruit cluster size $L$, we define a bounding box with size $b$ in the image space for ray casting boundaries:

\vspace{-0.4cm} 
\begin{equation}
\Scale[1.2]
{
    b = \frac{L * F_x}{z}
    \label{eq:bounding_region}
}
\end{equation}
\vspace{-0.5cm}

%
\noindent\textbf{(iv) Planning and Control:}
Given a set of best viewpoints for the current scene, we compute the order of execution following the traveling salesman planning algorithm. The MoveIt ROS package is then used to execute these viewpoints following collision-free trajectories.

\vspace{-0.2cm}
\section{Experiments}

Our approach is evaluated in Gazebo simulation using 8 sweet pepper plant models from \cite{zaenker2021viewpoint}. The three semantic classes used in all simulation experiments consist of the background, leaves, and the target semantic class, fruit. The experiments run on a laptop with an Intel Core i7-11800H (2.30 GHz) and 16 GB RAM.

\noindent\textbf{Metrics}
We evaluate the evolution of multi-class entropy and surface coverage defined as in \cite{asgharivaskasi2023semantic} and \cite{delmerico2018comparison} respectively, only for fruits as they are the main targets. While surface coverage is an indicator of reconstruction completeness, the entropy serves as a measure of uncertainty in this reconstruction. We compute the total entropy as the sum of entropy values (equation \ref{eq:h_x}) for all voxels inside a 3D bounding box enclosing each fruit cluster. The surface coverage $SC$ is computed as follows:

\vspace{-0.4cm} 
\begin{equation}
\Scale[1.1]
{
    SC = \frac{\text{Observed surface points}}{\text{Total points in ground truth model}}
    \label{eq:surface-coverage}
}
\end{equation}
\vspace{-0.4cm} 

A surface point in the ground truth model is considered observed if the closest point in the reconstruction point cloud is within a distance threshold equal to the map resolution. See some evaluation parameters in Table \ref{tab:parameters}. Additional parameters can be found in the supplementary material.



\begin{table}[htbp]
\centering
\caption{Parameters used during evaluation}
\vspace{-0.2cm}
\resizebox{0.5\textwidth}{!}{
\begin{tabular}{c|c|c|l}
    \toprule
    \textbf{Category} & \textbf{Parameter} & \textbf{Value} & \textbf{Description} \\
    \midrule
    \textbf{Mapping} & $\delta S$ [m] & 0.015 & Map resolution \\
    & $max\_range$ [m] & 1.0 & Max depth range for mapping \\
    \midrule
    \textbf{Segmentation} & $P_{gt}$ & 0.7 & Probability of correct classification \\
    {} & {} & {} & during semantic segmentation \\
    \midrule
    \textbf{Downsampling} & $L$ [m] & 0.1 & Typical size of a fruit cluster\\ 
    \midrule
    \textbf{NBV-planner} & $r$ [m] & 0.4 & Radius of sphere for viewpoint sampling \\
    & $N_{\phi}$ & 10 & Number of azimuth samples \\
    & $N_{\theta}$ & 5 & Number of elevation samples \\
    \midrule
    \textbf{OSAMCEP} & $max\_dist$ [m] & 0.1 & Max distance from the semantic  \\
    {} & {} & {} & target for relevant voxels \\
    \midrule
    \textbf{DBSCAN} & $\epsilon$ [m] & 0.05 & Radius of a neighborhood \\
    \bottomrule 
\end{tabular}
}
\label{tab:parameters}
\vspace{-0.6cm}
\end{table}



In addition, we simulate segmentation noise to account for the limitations of common segmentation models, especially in agricultural environments \cite{kootstra2024advances}. To this end, for every mask we assign the ground truth class with probability $P_{gt}$ and a wrong class with probability $(1-P_{gt})$.

\vspace{-0.2cm}
\subsection{Overall performance}
\vspace{-0.1cm}
In this part, we aim to evaluate our whole pipeline for active mapping using a 6 DOF UR3 manipulator on a Clearpath Husky robot \cite{husky_UR3_simulator} and an RGBD camera on the end-effector. We create a simulation environment with 8 sweet pepper plants distributed in rows as shown in Fig. \ref{reconstruction}. The mobile robot is given a sequence of waypoints to take the manipulator close to each pair of plants. At each position, the manipulator starts the initial scanning with 12 predefined and sparse viewpoints. Then, the next best viewpoints for the fruit clusters near the robot are computed and executed. Since some viewpoints are not reachable, we sample, evaluate, and execute new viewpoints if necessary until successfully executing 12, before moving to the next plants. We compare our approach with two baselines: (i) the frontier-based approach proposed by Zaenker \textit{et al.} \cite{zaenker2021viewpoint} described in section \ref{sec:relatedworks}. It is adapted to start with the initial predefined scanning as our method to ensure the same initialization, followed by 12 additional NVB successful viewpoint executions based on ROI frontiers as in the original implementation; (ii) a dense scanning approach consisting of the same number of viewpoints (24) for fair comparison. The results of 10 trials with different random seeds are averaged for each method.

\begin{figure}[htbp]
\centering
\includegraphics[width=80mm]{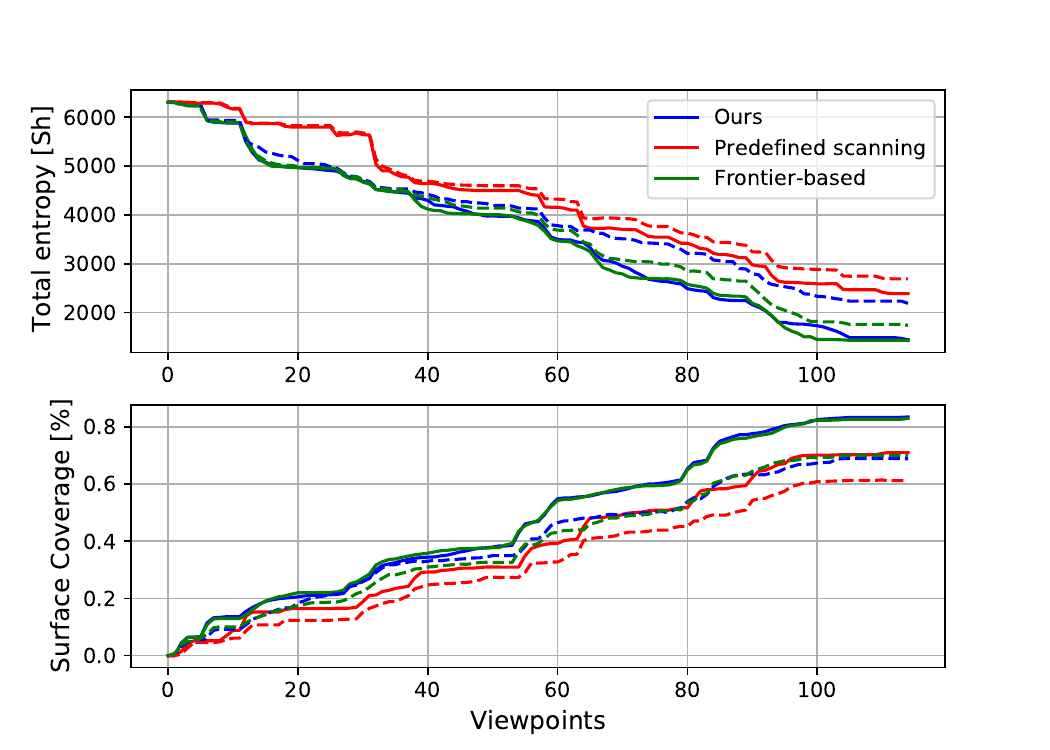} 
\caption{Total entropy and surface coverage with our active mapping approach, a frontier-based approach, and predefined scanning, with (dashed lines) and without (solid lines) segmentation noise. All the methods execute 120 viewpoints along the path between two crop rows with 8 plants in total. }
\label{fig:predefined_scanning_vs_ours}
\vspace{-0.5cm} 
\end{figure}

\begin{figure}[htbp]
\centering
\includegraphics[width=60mm]{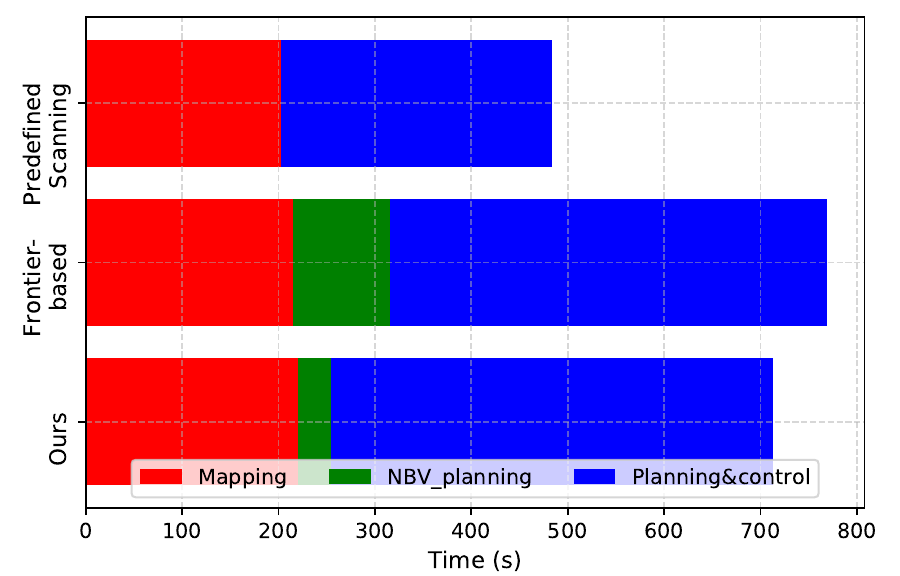} 
\caption{Runtime comparison between our approach and two baselines. The total time is divided into three tasks: mapping, NBV planning, and motion planning and control. Note that our approach significantly reduces the NBV planning time compared to the frontier-based method.}
\label{fig:runtime}
\vspace{-10pt} 
\end{figure}

Fig. \ref{fig:predefined_scanning_vs_ours} depicts the evolution of entropy and surface coverage as the robot navigates through the crop row, collecting views. Fig. \ref{fig:runtime} shows the corresponding runtime for each method. It is evident that the two active mapping approaches result in higher fruit surface coverage (approximately 11\% higher by the end of the experiment) and lower entropy compared to the predefined scanning method. This improvement can be attributed to the richer viewpoints computed by these methods, as both are information-based and target-aware. However, this enhancement comes at a cost, as the runtime for the active methods is more than 45\% higher than that of the predefined scanning approach (Fig. \ref{fig:runtime}). Although all three methods execute the same number of successful viewpoints, the active mapping approaches require the evaluation of the utility of each candidate viewpoint. Additionally, some of the best viewpoints are not reachable by the manipulator arm, resulting in increased computation time for motion planning and control.

\vspace{-0.4cm}


Between the two active mapping methods, the frontier-based approach and our cluster-centric method show similar performance in surface coverage and entropy. However, our method improves runtime by 8\% due to a more efficient NBV planning strategy. Specifically, we observed that most of the random viewpoint candidates generated by the frontier-based method are not reachable by the robot arm. Consequently, multiple rounds of sampling and evaluation are required to achieve successful viewpoints, which increases the NBV planning time.This issue becomes more pronounced over time, as the number of ROI frontier nodes decreases, affecting the diversity of viewpoint candidates.  In contrast, our approach uniformly samples candidate viewpoints from fruit cluster centroids, ensuring diversity and complete coverage.

It is worth noting in Fig. \ref{fig:predefined_scanning_vs_ours} that both surface coverage and entropy are significantly impacted by segmentation noise (indicated by dashed lines). Specifically, segmentation noise results in a 17\% reduction in surface coverage by the end of the experiment, which could pose a critical limitation in real-world applications. Finally, Fig. \ref{reconstruction} shows that the final reconstruction built by our active mapping approach is more complete than the one of the predefined scanning method. The baseline exhibits several blank centroids, which requires additional viewpoints for a complete reconstruction.


\vspace{-0.1cm}
\subsection{Ablation studies}
To minimize the influence of the constrained manipulator workspace, control and motion planning algorithm accuracy on the comparison of our active mapping approach with other mapping algorithms, we perform ablation studies using a free-moving camera. We separately assess the performance of our downsampling strategy for ray casting and the information utility function. Using 6 plant models, we run the active mapping algorithms for 30 viewpoints per plant across 10 trials with varying initializations. We then compute the average entropy and surface coverage across all plants.

\begin{figure}[htbp]
\centering
\includegraphics[width=80mm]{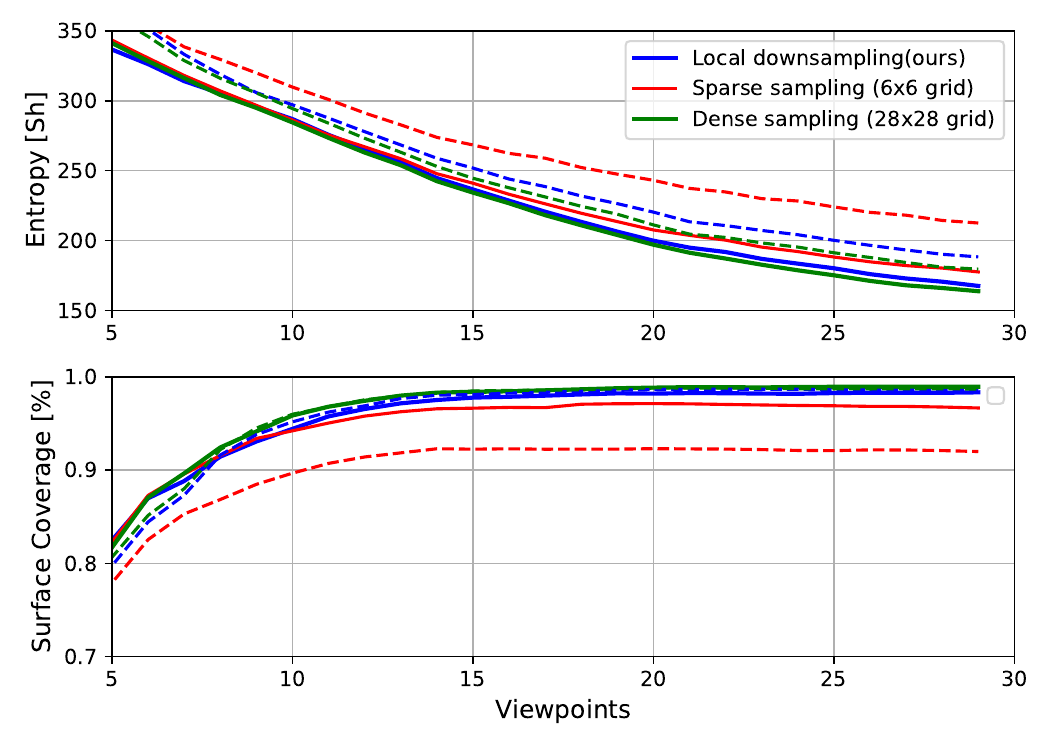} 
\caption{Entropy and surface coverage for different downsampling strategies. Solid lines and dashed lines indicate viewpoints taken at 0.4 m and 0.6 m from cluster centroids, respectively. Note that despite the distance from the targets, our downsampling strategy maintains close performance to dense sampling.}
\label{fig:downsampling}
\vspace{-0.4cm} 
\end{figure}

\vspace{2pt}
\noindent\textbf{Evaluation of our downsampling strategy for ray casting} The downsampling strategy in Section \ref{sec:methods} aims to accelerate ray casting while preserving viewpoint quality. We compared its performance with two baselines: dense sampling (28x28 grid) and sparse sampling (6x6 grid) from \cite{zaenker2021viewpoint}. Both baselines perform uniform sampling across the images. Since our downsampling approach is dependent on the distance between cluster centroids and viewpoints, we tested two distance values, 0.4m and 0.6m. Fig. \ref{fig:downsampling} shows our downsampling method achieves comparable performance to the dense sampling baseline by prioritizing image regions corresponding to semantic targets. Note that the sparse sampling baseline resulted in higher entropy and lower surface coverage over time, especially for larger distance values. This outcome is expected, as uniform sampling across the entire image can miss semantic areas depending on the distance from the scene. Our strategy mitigates this by parameterizing the downsampling process based on distance from the clusters and expected cluster size. The average ray tracing times were 2.58 ms for our method, 45.21 ms for dense sampling, and 2.50 ms for sparse sampling, indicating that our approach retains the efficiency of sparse downsampling while maintaining the viewpoint quality of dense sampling.

\begin{figure}[htbp]
\centering
\includegraphics[width=75mm]{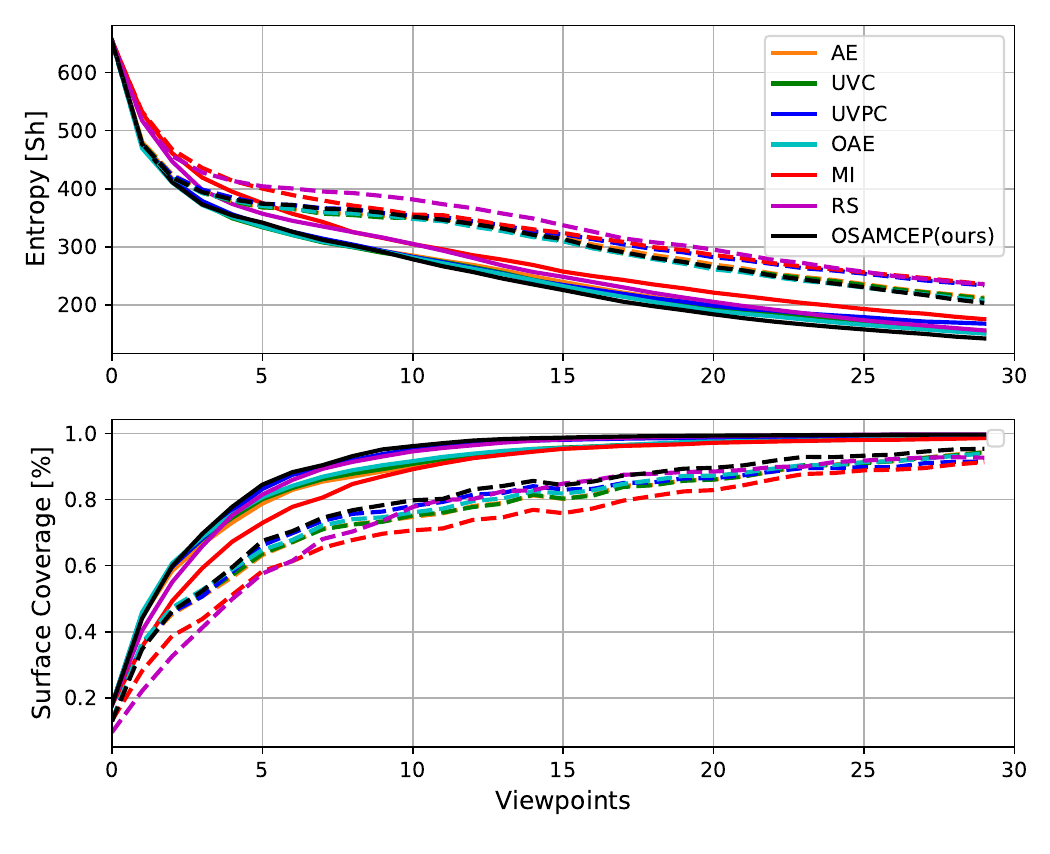} 
\caption{Entropy and surface coverage averaged over six plants vs the number of executed viewpoints for different information gain metrics (RS: Random sampling; AE: Average entropy; UVC: Unknown voxels count; UVPC: Unknown voxels with proximity count; OAE: Occlusion aware entropy; MI: Mutual information;  OSAMCEP: Occlusion and semantic Aware Multi-Class Entropy with Proximity
Count), with (dashed lines) and without (solid lines) segmentation noise.}
\label{fig:IG_metrics}
\vspace{-0.5cm} 
\end{figure}

\noindent\textbf{Evaluation of the Information Gain Metric}
We consider several information metrics as baselines, including Average Entropy (AE) \cite{delmerico2018comparison}, Unknown Voxels Count (UVC) \cite{delmerico2018comparison,zaenker2021viewpoint}, Unknown Voxels with Proximity Count \cite{zaenker2021viewpoint}, Occlusion Aware Entropy (OAE) \cite{delmerico2018comparison}, and Mutual Information (MI) \cite{asgharivaskasi2023semantic}.
We also consider the case of random sampling (RS).
Figure \ref{fig:IG_metrics} presents the results averaged across six plants, demonstrating that our proposed IG metric outperforms all baselines in reducing scene entropy and maximizing surface coverage with the fewest viewpoints. Interestingly, the majority of the IG metrics perform closely to our metric when segmentation noise is not considered, with even the random sampling strategy yielding competitive results. This suggests that active mapping can be effectively performed as long as the viewpoints are directed toward target clusters. This observation aligns with the findings reported in \cite{zaenker2021viewpoint}, where no significant difference in fruit coverage was observed when using different metrics. However, when segmentation noise is introduced, our utility function shows clear advantages over other metrics. For instance, to achieve 80\% of surface coverage, our metric requires an average of 10 viewpoints per plant, whereas other metrics require between 11 and 17 viewpoints. These results would translate into substantial efficiency gains in real-world mapping scenarios.
\vspace{-0.5cm} 

\subsection{Real-world experiments}

We perform real-world experiments with a custom 6-DOF robotic arm mounted on a wheeled platform. An Intel RealSense D405 camera (400x400 resolution) is attached to the end effector for RGBD data acquisition. The experiments take place in an urban high tunnel with tomato plants, using AprilTag markers \cite{olson2011tags} mounted on the ceiling for global localization.

Our mapping approach involves an initial predefined scanning phase, followed by NBV planning. The semantic classes of interest are fruits and background, with semantic segmentation performed using the Yolact model \cite{bolya2019yolact}.

\begin{figure}[htbp]
\centerline{\includegraphics[width=85mm]{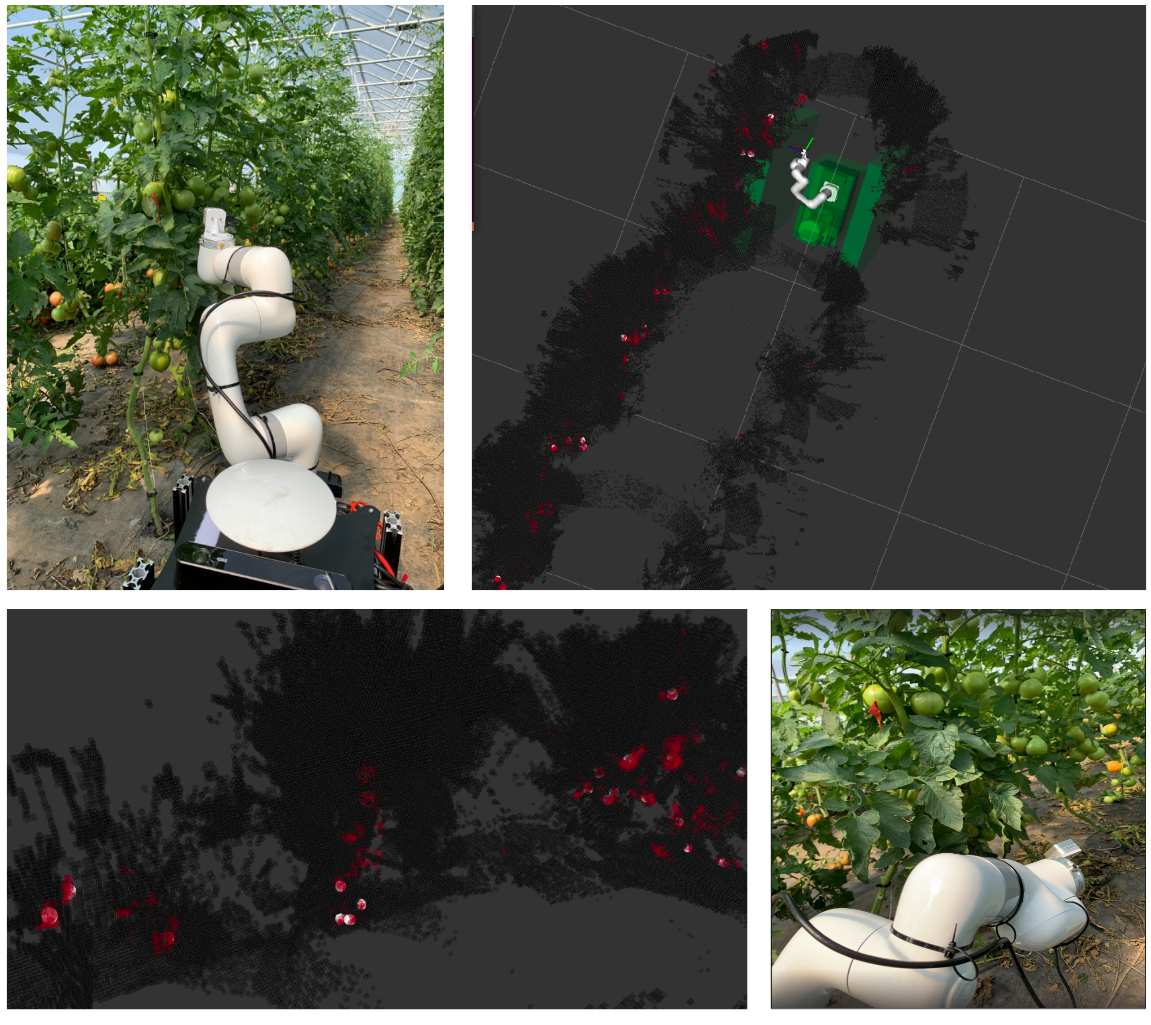}}
\caption{Top left: Mobile manipulator in a high tunnel with tomato plants. Top right: Reconstructed scene along a tomato row. Bottom left: An amplified view of the reconstruction (white spheres are cluster centroids; red voxels are fruit nodes; black voxels are background). Bottom right: Sample best viewpoints, in which the camera looks upwards to reveal fruit areas. }
\label{fig:mapping_hightunnel}
\vspace{-0.3cm}
\end{figure}

Fig. \ref{fig:mapping_hightunnel} shows the reconstructed tomato row scene. Our method effectively identifies complex viewpoints, exposing typically obscured fruit areas in fixed camera setups. However, challenges such as noisy depth maps caused by illumination variations, erratic plant motion due to variable wind speeds that violate the static-environment assumption critical to our approach, and segmentation errors, including missed detections or incorrect label assignments from crop variability, reduce reconstruction accuracy, as seen in the \href{https://youtu.be/l6IBl4n1GJ0}{supplementary video}.

\vspace{-0.1cm}

\vspace{-0.2cm}
\section{Conclusion}
This paper presented an approach for target-aware active semantic mapping in horticultural environments. By leveraging semantic octrees, semantic targets were effectively found. We introduced a cluster-centric approach for NBV planning which enhances surface coverage and reduces the entropy of the reconstructions. In addition, a downsampling strategy was proposed to improve the efficiency of ray casting, achieving significant improvements compared to traditional methods. Finally, we presented an information utility function that considers visibility, proximity, and semantics to evaluate viewpoints, outperforming other metrics especially when segmentation noise is considered.

However, some limitations still remain. For example, the accuracy of the geometric reconstruction depends significantly on the map resolution, which directly impacts the computational demand. A future direction to address this issue could be combining our NBV planning method with state-of-the-art reconstruction methods (e.g. NeRF, or Gaussian Splatting) to achieve a more accurate reconstruction of agricultural scenes.

\bibliographystyle{IEEEtran}
\bibliography{references}
\end{document}